\documentclass{article}

\usepackage[nonatbib, preprint]{nips_2018}

\usepackage{times}
\usepackage{cite}
\usepackage{subfigure}
\usepackage{graphicx}
\usepackage{amsmath}
\usepackage{bm}
\usepackage{url}
\usepackage{stmaryrd}
\usepackage{balance}
\usepackage{amssymb}
\usepackage{pgfplots}
\usepackage{pifont}
\usepackage[usenames,dvipsnames]{pstricks}
\usepackage{epsfig}
\usepackage{booktabs}
\usepackage{pgffor}
\usepackage{tikz}
\usepackage{multirow}
\usepackage{bbm}

\newcommand*{\QEDB}{\hfill\ensuremath{\square}}

\newcommand{\mb}{\mathbf}

\newtheorem{theo}{\textsc{Theorem}}

\newtheorem{proof}{\textsc{Proof}}

\newcommand{\our}{\textsc{Gadam}}

\usepackage{algorithm}
\usepackage{algorithmic}

\title{GADAM: Genetic-Evolutionary ADAM for Deep Neural Network Optimization}

\author{Jiawei~Zhang$^\star$, Fisher B. Gouza$^\star$\\
$^\star$IFM Lab, Florida State University, FL, USA\\
jiawei@ifmlab.org, fisherbgouza@gmail.com}

\begin{document}

\maketitle

\vspace{-20pt}
%------------------------------------------

\begin{abstract}
\vspace{-10pt}
Deep neural network learning can be formulated as a non-convex optimization problem. Existing optimization algorithms, e.g., Adam, can learn the models fast, but may get stuck in local optima easily. In this paper, we introduce a novel optimization algorithm, namely {\our} (Genetic-Evolutionary Adam). {\our} learns deep neural network models based on a number of unit models generations by generations: it trains the unit models with Adam, and evolves them to the new generations with genetic algorithm. We will show that {\our} can effectively jump out of the local optima in the learning process to obtain better solutions, and prove that {\our} can also achieve a very fast convergence. Extensive experiments have been done on various benchmark datasets, and the learning results will demonstrate the effectiveness and efficiency of the {\our} algorithm.

\end{abstract}

%------------------------------------------
\vspace{-20pt}
\section{Introduction}\label{sec:intro}
\vspace{-10pt}

Adam (Adaptive Momentum Estimation) \cite{KB14}, with sound theoretic foundations, has been widely used as the optimization algorithm in many existing machine learning models, especially the deep neural networks with non-convex objective functions. Formally, learning the variables of deep neural network models can be formally represented as the following optimization problem:\vspace{-3pt}
\begin{equation}
\min_{\mb{w} \in \mathcal{W}} f(\mb{X}, \mb{y}; \mb{w}),
\end{equation}
where $f(\cdot, \cdot ; \mb{w})$ denotes the loss function parameterized by variable vector $\mb{w} \in \mathcal{W}$ and $\mb{X}$, $\mb{y}$ denote the features and labels of the data instances respectively. Set $\mathcal{W}$ represents the feasible solution space, which can be $\mathbb{R}^n$ for simplicity if there exist no other constraints on the variables.

Adam adopts an iterative learning scheme to update the model variables. In iteration $k$, given a training instance $(\mb{x}_i, y_i)$ from the training set, the updating equation of variable $\mb{w}$ can be represented as \vspace{-5pt} \begingroup\makeatletter\def\f@size{8.5}\check@mathfonts
\begin{alignat}{2}\label{eq:adam}
\mb{w}_{k} &= \mb{w}_{k-1} - \eta_k  \frac{ \sqrt{ 1 - \beta^k_2 } }{1 - \beta^k_1} \frac{ \mb{m}_{k-1} }{ \sqrt{\mb{v}_{k-1}} + \epsilon},
\mbox{ where}
\begin{cases}
\mb{m}_{k-1} &= \beta_1 \mb{m}_{k-2} + (1-\beta_1) \nabla f(\mb{x}_i, {y}_i; \mb{w}_{k-1});\\
\mb{v}_{k-1} &= \beta_2 \mb{v}_{k-2} + (1-\beta_2) \nabla f(\mb{x}_i, {y}_i; \mb{w}_{k-1})^2.
\end{cases}
\end{alignat}\endgroup
where $\mb{w}_{k}$ represents the variable in the $k_{th}$ iteration and $\nabla f(\mb{x}_i, y_i; \mb{w}_{k-1})$ denotes the computed gradient of the function regarding variable $\mb{w}$ based on the input training instance. Term $\eta_k$ is the learning rate, which is usually a small constant. Moment vectors $\mb{m}_{k_1}$ and $\mb{v}_{k-1}$ will be adapted in the learning process, and $\beta_1$ and $\beta_2$ are the corresponding momentum parameters.

Adam together with its many variants \cite{D16, KS17} have been shown to be effective for optimizing a large group of problems. However, for the non-convex objective functions of deep learning models, Adam cannot guarantee to identify the globally optimal solutions, whose iterative updating process may inevitably get stuck in local optima. The performance of Adam is not very robust, which will be greatly degraded for the objective function with non-smooth shape or learning scenarios polluted by noisy data. Furthermore, the distributed computation process of Adam requires heavy synchronization, which may hinder its adoption in large-cluster based distributed computational platforms. 

On the other hand, genetic algorithm (GA), a metaheuristic algorithm inspired by the process of natural selection in evolutionary algorithms, has also been widely used for learning the solutions of many optimization problems. In GA, a population of candidate solutions will be initialized and evolved towards better ones. Several attempts have also been made to use GA for training deep neural network models \cite{ZCG18, SMCLSC17, DG17} instead of the gradient descent based methods. GA has demonstrated its outstanding performance in many learning scenarios, like {non-convex objective function containing multiple local optima}, {objective function with non-smooth shape}, as well as {a large number of parameters} and {noisy environments}. GA also fits the parallel/distributed computing setting very well, whose learning process can be easily deployed on parallel/distributed computing platforms. Meanwhile, compared with Adam, GA may take more rounds to converge in addressing optimization objective functions.

In this paper, we propose a new optimization algorithm, namely {\our} (Genetic Adaptive Momentum Estimation), which incorporates Adam and GA into a unified learning scheme. Instead of learning one single model solution, {\our} works with a group of potential unit model solutions. In the learning process, {\our} learns the unit models with Adam and evolves them to the new generations with genetic algorithm. In addition, Algorithm {\our} can work in both standalone and parallel/distributed modes, which will also be studied in this paper.

The following part of the paper is organized as follows. In Section~\ref{sec:relatedwork}, we will introduce the related research works about Adam and GA. The detailed learning algorithm {\our} will be introduced in Section~\ref{sec:method}, whose performance analysis is available in Section~\ref{sec:analysis}. In Section~\ref{sec:experiment}, we will provide the numerical experiments of {\our} compared with the existing learning algorithms in training several representative deep neural network models. Finally, we will conclude this paper in Section~\ref{sec:conclusion}.

%Based on the pre-specified candidate solution fitness functions, the learning process of GA involves \textit{initialization}, \textit{crossover}, \textit{mutation} and \textit{selection}. More specifically, the population \textit{initialization} operation will generate a group of initial solutions, from which the GA algorithm will ``search'' for better candidate solutions. The search process is achieved with the \textit{crossover} operation, which allows GA to evolve the current candidate solutions to new candidate solutions. Meanwhile, to avoid sinking into the local optimal points, GA adopts \textit{mutation} to jump out from them. The \textit{selection} operation based on the fitness function picks good candidate solutions for the next round of search process.
%------------------------------------------
\vspace{-10pt}
\section{Related Works} \label{sec:relatedwork}
\vspace{-10pt}

\noindent \textbf{Optimization Algorithms}: In addressing the optimization problems, SGD (Stochastic Gradient Descent) \cite{RM51} with sound theoretic foundations has been one of the most frequently used algorithms. Given an objective function $\min_{\mb{w} \in \mathcal{W}} f(\mb{w})$, SGD adopts an iterative updating schema to learn the model variables as follows:
\begin{equation}
\mb{w}_k = \mb{w}_{k-1} - \eta_{k-1} \cdot \nabla f(\mb{w}_{k-1}).
\end{equation}
Here, the parameter $\eta_{k}$ is usually a small constant, which is referred to as the learning rate. Meanwhile, to further accelerate the learning process, a variant of SGD, namely SGDM (SGD with Momentum) \cite{SMDH13}, has been proposed, which updates the variables according to the following equations
\begin{equation}
\mb{w}_k = \mb{w}_{k-1} - \eta_{k} \cdot \mb{v}_k \mbox{, where } \mb{v}_k = \beta \cdot \mb{v}_{k-1} + \nabla f(\mb{w}_{k-1}).
\end{equation}
In the equation, $\beta \in [0, 1)$ is a momentum parameter and vector $\mb{v}_0$ is initialized with value $\mb{0}$.

SGD and SGDM scale the gradient uniformly in all directions (i.e., all the variables), which makes the tuning of learning rate $\eta_{k}$ very tedious and laborious. To resolve such a problem, several adaptive optimization methods have been proposed, including Adam \cite{KB14}, RMSprop \cite{Tieleman2012} and Adagrad \cite{DHS11}, which uses distinct learning rates for different variables. Formally, the variable updating equations adopted in Adagrad \cite{DHS11} and RMSprop \cite{Tieleman2012} can be represented as follows respectively:
\begin{alignat}{2}
\begin{cases}
&\mbox{\# Adagrad}\\
&\mb{w}_{k} = \mb{w}_{k-1} - \eta_{k} \cdot \frac{ \nabla f(\mb{w}_{k-1}) }{\sqrt{\mb{v}_{k-1}} + \epsilon},\\
&\mb{v}_{k-1} = \sum_{j = 1}^{k-1} \nabla f(\mb{w}_{j})^2.
\end{cases}
\mbox{ }
\begin{cases}
&\mbox{\# RMSprop}\\
&\mb{w}_{k} = \mb{w}_{k-1} - \eta_{k} \cdot \frac{ \nabla f(\mb{w}_{k-1}) }{\sqrt{\mb{v}_{k-1}} + \epsilon},\\
&\mb{v}_{k-1} = \beta \cdot \mb{v}_{k-2} + (1-\beta) \cdot \nabla f(\mb{w}_{k-1})^2.
\end{cases}
\end{alignat}
In Adagrad, vector $\mb{v}_{k}$ increases monotonically as the update iteration continues, and its scaling factor will keep decreasing when updating vector $\mb{w}_{k}$. RMSprop addresses the problem by employing an average scale instead of a cumulative scale, and Adam \cite{KB14} resolves such a problem with a bias correction, which adopts an exponential moving average for the step in lieu of the gradient. 

%Compared with SGD, Adam has been shown to work well in the initial learning iterations, but it has been found to generalize poorly compared with SGD. In a recent work \cite{KS17}, the authors propose an approach, which is capable to switch between Adam and SGD in the learning process.

\noindent \textbf{Genetic Algorithms}: In the 50s in the last century, computer scientists proposed that evolutionary systems with the idea that evolution could be used as an optimization tool for engineering problems. The central idea in all these systems was to evolve a population of candidate solutions to a given problem with the operations inspired by the natural genetic variation and selection. Genetic algorithm \cite{H92} invented by John Holland is one of the most important algorithms proposed in the evolutionary computing. A detailed introduction to classic genetic algorithm is available in \cite{M98}. In recent years, some research works proposed to apply genetic to evolve neural network models. Eli et al. propose to extend previous work and propose a genetic algorithm assisted method for deep learning \cite{OI14, SMCLSC17}. In \cite{ZCG18}, Zhang et al. introduce a evolutionary network model as an alternative approach to deep learning models based on data sampling, genetic model learning and ensemble learning.

\noindent \textbf{Deep Learning}: The essence of deep learning is to compute hierarchical features or representations of the observational data \cite{GBC16, LBH15}. With the surge of deep learning research and applications in recent years, lots of research works have appeared to apply the deep learning methods, like deep belief network \cite{HOT06}, deep Boltzmann machine \cite{SH09}, deep neural network \cite{J02, KSH12} and deep autoencoder model \cite{VLLBM10}, in various applications. The representative deep neural network models studied nowadays include the convolutional neural network (CNN) \cite{726791, KSH12}, recurrent neural network (RNN), long-short term memory (LSTM) \cite{HS97}, deep feedforward neural network \cite{GB10}, and deep auto-encoder model \cite{VLLBM10}. These deep neural network models have achieved remarkable performance in various application domains, like speech and audio processing \cite{DHK13, HDYDMJSVNSK12}, language processing \cite{ASKR12, MH09}, information retrieval \cite{H12, SH09}, computer vision \cite{LBH15}, as well as multimodal and multi-task learning \cite{WBU10, WBU11}.

% In this paper, we will use deep learning models as an example to illustrate the {\our} algorithm.

%------------------------------------------
\vspace{-10pt}
\section{Proposed Methods}\label{sec:method}
\vspace{-10pt}

%----------------
\begin{figure*}[!t]
 \centering    
 \begin{minipage}[l]{1.0\columnwidth}
  \centering
    \includegraphics[width=1.0\textwidth]{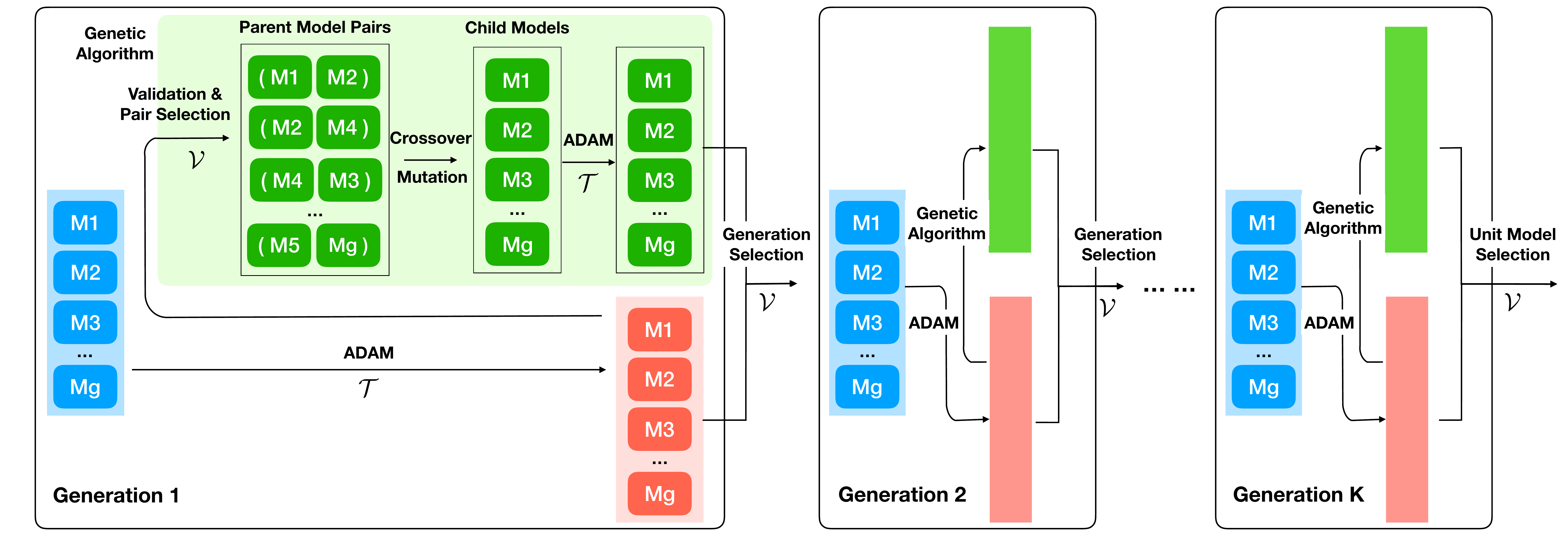}
 \end{minipage}
\caption{Overall Architecture of {\our} Model.}\label{fig:framework}\vspace{-15pt}
\end{figure*}
%-----------------

The overall framework of the model learning process in {\our} is illustrated in Figure~\ref{fig:framework}, which involves multiple learning generations. In each generation, a group of unit model variable will be learned with Adam from the data, which will also get evolved effectively via the genetic algorithm. Good candidate variables will be selected to form the next generation. Such an iterative model learning process continues until convergence, and the optimal unit model variable at the final generation will be selected as the output model solution. For simplicity, in this paper, we will refer to unit models and their variables interchangeably without distinguishing their differences.

\vspace{-10pt}
\subsection{Model Population Initialization}
\vspace{-8pt}

{\our} learns the optimal model variables based on a set of unit models (i.e., variables of these unit models by default), namely the unit model population, where the initial unit model generation can be denoted as set $\mathcal{G}^{(0)} = \{M^{(0)}_1, M^{(0)}_2, \cdots, M^{(0)}_g\}$ ($g$ is the population size and the superscript represents the generation index). Based on the initial generation, {\our} will evolve the unit models to new generations, which can be represented as $\mathcal{G}^{(1)}, \mathcal{G}^{(2)}, \cdots, \mathcal{G}^{(K)}$ respectively. Here, parameter $K$ denotes the total generation number.

For each unit model in the initial generation $\mathcal{G}^{(0)}$, e.g., $M^{(0)}_i$, we can initialize its variables $\mb{w}^{(0)}_i$ with random values sampled from certain distributions (e.g., the standard normal distribution). These initial generation serve as the starting search points, from which {\our} will expand to other regions to identify the optimal solutions. Meanwhile, for the unit models in the following generations, their variable values will be generated via GA from their parent models respectively. 

\vspace{-10pt}
\subsection{Model Learning with Adam}
\vspace{-8pt}

In the learning process, given any model generation $\mathcal{G}^{(k)}$ ($k \in \{1, 2, \cdots, K\}$), {\our} will learn the (locally) optimal variables for each unit model with Adam. Formally, {\our} trains the unit models with several epochs. In each epoch, for each unit model $M^{(k)}_i \in \mathcal{G}^{(k)}$, a separated training batch will be randomly sampled from the training dataset, which can be denoted as $\mathcal{B} = \{(\mb{x}_1, \mb{y}_1), (\mb{x}_2, \mb{y}_2), \cdots, (\mb{x}_b, \mb{y}_b)\} \subset \mathcal{T}$ (here, $b$ denotes the batch size and $\mathcal{T}$ represents the complete training set). Let the loss function introduced by unit model $M^{(k)}_i$ for instance $(\mb{x}_j, \mb{y}_j)$ be $f(\mb{x}_j, \mb{y}_j; \mb{w}_i^{(k)})$, and we can represent the learned model variable vector by the training instance as
\begin{equation}
\bar{\mb{w}}_i^{(k)} = Adam\left( f(\mb{x}_j, \mb{y}_j; \mb{w}_i^{(k)}); \mathcal{B} \right).
\end{equation}
Depending on the specific unit models and application settings, the loss function will have different representations, e.g., mean square loss, hinge loss or cross entropy loss. In this paper, we will illustrate {\our} as a general learning algorithm without providing the concrete loss representations. Such a learning process continues until convergence, and we can represent the updated model generation $\mathcal{G}^{(k)}$ with (locally) optimal variables as $\bar{\mathcal{G}}^{(k)} = \{\bar{M}^{(k)}_1, \bar{M}^{(k)}_2, \cdots, \bar{M}^{(k)}_g\}$, whose corresponding variable vectors can be denoted as $\{\bar{\mb{w}}^{(k)}_1, \bar{\mb{w}}^{(k)}_2, \cdots, \bar{\mb{w}}^{(k)}_g\}$ respectively.

\vspace{-10pt}
\subsection{Model Evolution with Genetic Algorithm}
\vspace{-6pt}

In this part, based on the learned unit models, i.e., $\bar{\mathcal{G}}^{(k)}$, we propose to further search for better solutions for the unit models effectively via the genetic algorithm.

\vspace{-10pt}
\subsubsection{Model Fitness Evaluation}
\vspace{-6pt}

Unit models with good performance may fit the learning task better. Instead of evolving models randomly, {\our} proposes to pick good unit models from the current generation to evolve. Based on a sampled validation batch $\mathcal{V} \subset \mathcal{T}$, the fitness score of unit models, e.g., $\bar{M}^{(k)}_i$, can be effectively computed based on its loss terms introduced on $\mathcal{V}$ as follows
\begin{equation}
\mathcal{L}^{(k)}_i = \mathcal{L}(\bar{M}^{(k)}_i; \mathcal{V}) = \sum_{(\mb{x}_j, \mb{y}_j) \in \mathcal{V}} l(\mb{x}_j, \mb{y}_j; \bar{\mb{w}}_i^{(k)}).
\end{equation}
Based on the computed loss values, the selection probability of unit model $\bar{M}^{(k)}_i$ can be defined with the following softmax equation
\begin{equation}\label{equ:probability}
P(\bar{M}^{(k)}_i) = \frac{ \exp^{(- \hat{\mathcal{L}}^{(k)}_i ) }}{ \sum_{j = 1}^g \exp^{(- \hat{\mathcal{L}}^{(k)}_j) }}.
\end{equation}
Necessary normalization of the loss values is usually required in real-world applications, as $\exp^{(- {\mathcal{L}}^{(k)}_i )}$ may approach $0$ or $\infty$ for extremely large positive or small negative loss values ${\mathcal{L}}^{(k)}_i$. As indicated in the probability equation, the normalized loss terms of all unit models can be formally represented as $[\hat{\mathcal{L}}^{(k)}_1, \hat{\mathcal{L}}^{(k)}_2, \cdots, \hat{\mathcal{L}}^{(k)}_g]^\top$, where $\hat{\mathcal{L}}^{(k)}_i \in [0, 1], \forall i \in \{1, 2, \cdots, g\}$. According to the computed probabilities, from the unit model set $\bar{\mathcal{G}}^{(k)}$, $g$ pairs of unit models will be selected with replacement as the parent models for evolution, which can be denoted as $\mathcal{P} = \{ ( \bar{M}^{(k)}_{i_1}, \bar{M}^{(k)}_{j_1} ) , ( \bar{M}^{(k)}_{i_2}, \bar{M}^{(k)}_{j_2} ) , \cdots, ( \bar{M}^{(k)}_{i_g}, \bar{M}^{(k)}_{j_g} ) \}$.

\vspace{-10pt}
\subsubsection{Unit Model Crossover}
\vspace{-6pt}

Given a unit model pair, e.g., $( \bar{M}^{(k)}_{i_p}, \bar{M}^{(k)}_{j_p} ) \in \mathcal{P}$, {\our} inherits their variables to the child model via the \textit{crossover} operation. In crossover, the parent models, i.e., $\bar{M}^{(k)}_{i_p}$ and $\bar{M}^{(k)}_{j_p}$, will also compete with each other, where the parent model with better performance tend to have more advantages. We can represent the child model generated from $( \bar{M}^{(k)}_{i_p}, \bar{M}^{(k)}_{j_p} )$ as $\tilde{M}^{(k)}_{p}$. For each entry in the weight variable $\tilde{\mb{w}}^{(k)}_p$ of the child model $\tilde{M}^{(k)}_{p}$, {\our} initializes its values as follows:
\begin{alignat}{2}
\tilde{w}^{(k)}_p[m] = \mathbbm{1} (\mbox{rand} \le p_{i_p, j_p}^{(k)} ) \cdot \bar{w}^{(k)}_{i_p}[m] + \mathbbm{1} (\mbox{rand} > p_{i_p, j_p}^{(k)} )  \cdot \bar{w}^{(k)}_{j_p}[m].
\end{alignat}
In the equation, binary function $\mathbbm{1}(\cdot)$ returns $1$ iff the condition holds. Term ``$\mbox{rand}$'' denotes a random number in $[0, 1]$. The probability threshold $p_{i_p, j_p}^{(k)}$ is defined based on the parent models' performance:
\begin{equation}
p_{i_p, j_p}^{(k)} = \frac{ \exp^{ ( - \hat{\mathcal{L}}^{(k)}_{i_p} ) } }{ \exp^{ (- \hat{\mathcal{L}}^{(k)}_{i_p} ) } + \exp^{ (- \hat{\mathcal{L}}^{(k)}_{j_p}) } }.
\end{equation}
If $\hat{\mathcal{L}}^{(k)}_{i_p} > \hat{\mathcal{L}}^{(k)}_{j_p}$, i.e., model $\bar{M}_{i_p}^{(k)}$ introduces a larger loss than $\bar{M}_{j_p}^{(k)}$, we will have $0 < p_{i_p, j_p}^{(k)} < \frac{1}{2}$.

With such a process, based on the whole parent model pairs in set $\mathcal{P}$, we will be able to generate the children model set as set $\tilde{\mathcal{G}}^{(k)} = \{ \tilde{M}^{(k)}_{1}, \tilde{M}^{(k)}_{2}, \cdots, \tilde{M}^{(k)}_{g} \}$.

\vspace{-10pt}
\subsubsection{Unit Model Mutation}
\vspace{-6pt}

To avoid the unit models getting stuck in local optimal points, {\our} adopts an operation called \textit{mutation} to adjust variable values of the generated children models in set $\{ \tilde{M}^{(k)}_{1}, \tilde{M}^{(k)}_{2}, \cdots, \tilde{M}^{(k)}_{g} \}$. Formally, for each child model $\tilde{M}^{(k)}_{q}$ parameterized with vector $\tilde{\mb{w}}^{(k)}_{q}$, we propose to mutate the variable vector according to the following equation, where its $m_{th}$ entry can be updated as
\begin{alignat}{2}
\tilde{{w}}^{(k)}_{q}[m] = \mathbbm{1}(\mbox{rand} \le p_q) \cdot \mbox{rand}(0, 1) + \mathbbm{1}(\mbox{rand} > p) \cdot \tilde{{w}}^{(k)}_{q}[m],
\end{alignat}
In the equation, term $p_q$ denotes the \textit{mutation rate}, which is strongly correlated with the parent models' performance:
\begin{equation}
p_q = p \cdot \left(1 - P(\bar{M}^{(k)}_{i_q}) - P(\bar{M}^{(k)}_{j_q}) \right).
\end{equation}
For the child models with good parent models, they will have lower mutation rates. Term $p$ denotes the base mutation rate which is usually a small value, e.g., $0.01$, and probabilities $P(\bar{M}^{(k)}_{i_q})$ and $P(\bar{M}^{(k)}_{j_q})$ are defined in Equ~\ref{equ:probability}. These unit models will be further trained with Adam until convergence, which will lead to the $k_{th}$ children model generation $\tilde{\mathcal{G}}^{(k)} = \{ \tilde{M}^{(k)}_{1}, \tilde{M}^{(k)}_{2}, \cdots, \tilde{M}^{(k)}_{g} \}$.

\vspace{-10pt}
\subsection{New Generation Selection and Evolution Stop Criteria}\label{subsec:selection}
\vspace{-8pt}

Among these learned unit models in sets $\bar{\mathcal{G}}^{(k)} = \{\bar{M}^{(k)}_1, \bar{M}^{(k)}_2, \cdots, \bar{M}^{(k)}_g\}$ and $\tilde{\mathcal{G}}^{(k)} = \{\tilde{M}^{(k)}_1, \tilde{M}^{(k)}_2, \cdots, \tilde{M}^{(k)}_g\}$, we will re-evaluate their fitness scores based on a shared new validation batch. Among all the unit models in $\bar{\mathcal{G}}^{(k)} \cup \tilde{\mathcal{G}}^{(k)}$, the top $g$ unit models will be selected to form the $(k+1)_{th}$ generation, which can be formally represented as set $\mathcal{G}^{(k+1)} = \{M^{(k+1)}_1, M^{(k+1)}_2, \cdots, M^{(k+1)}_g \}$. Such an evolutionary learning process will stop if the maximum generation number has reached or there is no significant improvement between consequential generations, e.g., $\mathcal{G}^{(k)}$ and  $\mathcal{G}^{(k+1)}$: 
\begin{equation}\label{eq:convergence}
\Bigg | \sum_{M^{(k)}_i \in \mathcal{G}^{(k)}} \mathcal{L}^{(k)}_i - \sum_{M^{(k+1)}_i \in \mathcal{G}^{(k+1)}} \mathcal{L}^{(k+1)}_i \Bigg | \le \theta,
\end{equation}
The above equation defines the stop criterion of {\our}, where $\theta$ is the evolution stop threshold.

%------------------------------------------
\vspace{-10pt}
\section{Optimization Algorithm Analysis}\label{sec:analysis}
\vspace{-10pt}

%----------------
\begin{figure*}[!t]
 \centering    
 \begin{minipage}[l]{1.0\columnwidth}
  \centering
    \includegraphics[width=1.0\textwidth]{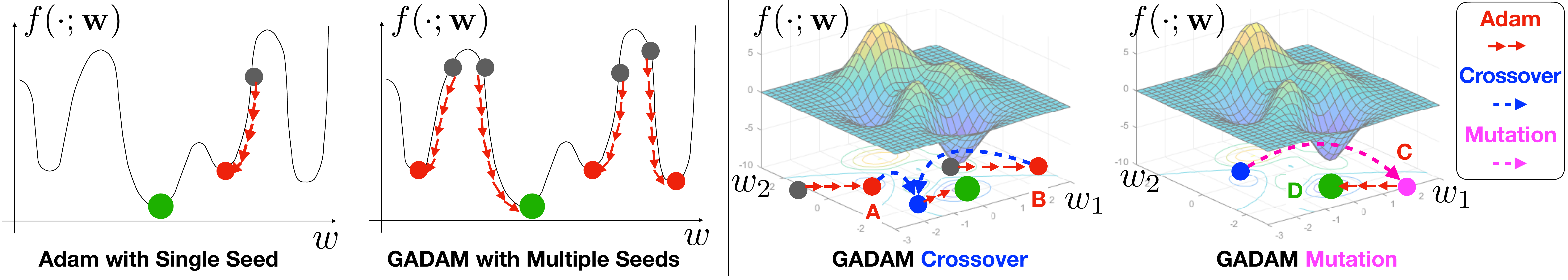}
 \end{minipage}
\caption{Analysis of {\our} Algorithm.}\label{fig:analysis}\vspace{-15pt}
\end{figure*}
%-----------------

In this section, we will analyze the optimization algorithm {\our} from the performance, convergence and parallel/distributed deployment perspectives respectively.

\vspace{-10pt}
\subsection{Learning Algorithm Performance Analysis}
\vspace{-8pt}

The optimization algorithm {\our} proposed in this paper actually incorporate the advantages of both Adam and genetic algorithm. With Adam, the unit models can effectively achieve the (locally/globally) optimal solutions very fast with a few training epochs. Meanwhile, via genetic algorithm based on a number of unit models, it will also provide the opportunity to search for solutions from multiple starting points and jump out from the local optima.

In Figure~\ref{fig:analysis}, we compare {\our} with Adam to illustrate its advantages. First of all, as shown in the left 2 plots in Figure~\ref{fig:analysis}, different from the traditional Adam algorithm with one single starting point, {\our} involves multiple starting points (i.e., multiple unit models) to search for the optima simultaneously. Both Adam and {\our} are based on the adaptive gradient descent to update the variables by following the function gradient. With more starting points, {\our} will have a larger chance to achieve the globally optimal solution, i.e., the green dot.

In the right two plots, we use a 2-variable function to illustrate that {\our} may jump out from the local optima via \textit{crossover} and \textit{mutation}. As shown in the plots, the loss function $f(\cdot, [w_1, w_2])$ is non-convex and has multiple local minima and one global minimum (i.e., the green dot). In the $3_{rd}$ plot, given two starting points (i.e., the grey dots), by following the gradient direction, {\our} will be able to update its variables and move to the local minima (i.e., dots $A$ and $B$). Via \textit{crossover}, a child model can be generated from the parent models, which is denoted by the blue dot in the plot. For the new child unit model, it inherits $w_1$ value from $A$ and $w_2$ value from $B$, which lies in a new region. By following the gradient from the new point, the global optimal point will be reached. Meanwhile, in the right plot, given a generated child model parameterized by the blue dot, via the \textit{mutation} operation, it will be able to jump out from the local minimal region to point $C$, from which the global optimum will be achieved.

%----------------
\begin{table*}[t]
\begin{minipage}[b]{0.45\linewidth}
\centering    
\scriptsize
 \begin{tabular}{| l | c c |} 
 \hline
 Optimization Algorithms & Avg. Loss & Accuracy Rate\% \\
  \hline\hline
{\our} &0.0396  &99.37   \\ 
 \hline
 Adam &0.0662 & 99.18  \\
 \hline
 SGD & 0.2166 & 94.51  \\
 \hline
 RMSProp & 0.0745 & 99.00  \\
 \hline
 AdaGrad & 0.1720 & 95.96  \\ 
 \hline
 AdaDelta & 1.8888 & 66.08  \\ 
 \hline \hline
 Existing Methods & Avg. Loss & Accuracy Rate\% \\
  \hline \hline
 LeNet-5 (GD) &N.A. &99.05 \cite{726791} \\
  \hline
 gcForest &N.A. &99.26 \cite{ijcai2017-497} \\
  \hline
 Deep Belief Net &N.A. &98.75 \cite{HOT06} \\
  \hline
 Random Forest &N.A. &96.8 \cite{ijcai2017-497} \\
  \hline
 SVM (rbf) &N.A. &98.60 \cite{DS02} \\
 \hline
\end{tabular}
\caption{Performance of Comparison Optimization Algorithms on the Testing Set.}\label{tab:mnist_result}
\end{minipage}\hfill   
\begin{minipage}[b]{0.45\linewidth}
  \centering
    \includegraphics[width=1.0\textwidth]{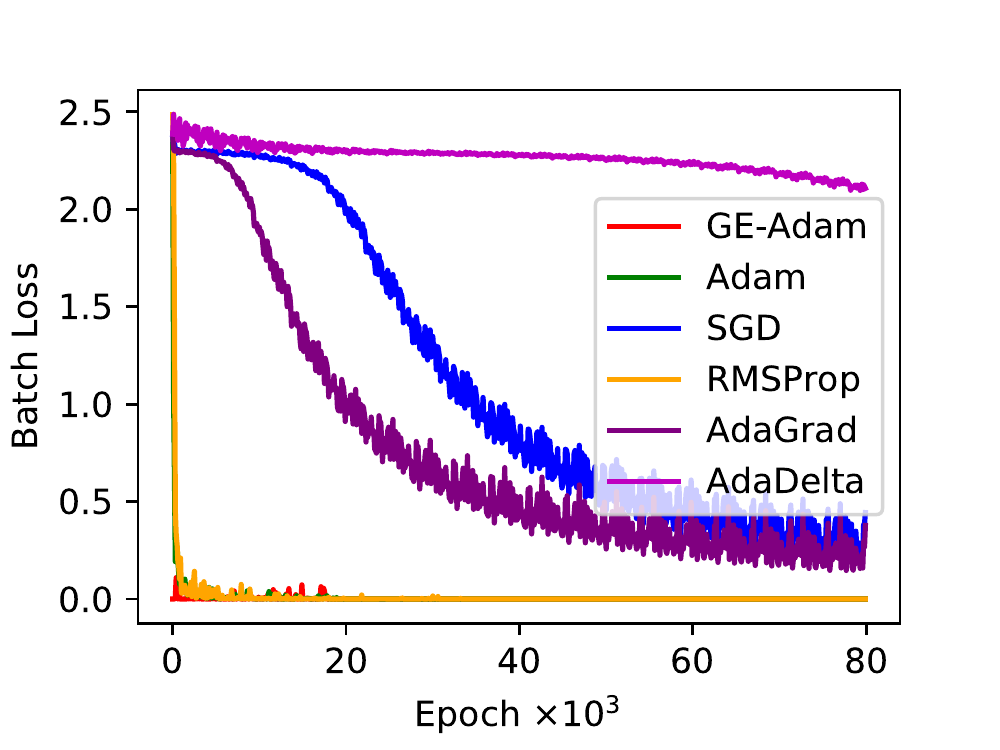}
 \caption{Training Loss of Comparison Optimization Algorithms.}\label{fig:train_loss}
 \end{minipage}
 \vspace{-15pt}
\end{table*}

\vspace{-10pt}
\subsection{Convergence Analysis}
\vspace{-8pt}

According to \cite{KB14}, for a smooth function, Adam will converge as the function gradient vanishes. On the other hand, the genetic algorithm can also converge according to \cite{TGDSM94}. Based on these prior knowledge, we can prove the convergence of {\our} as follows.

\vspace{-5pt}
\begin{theo}
Model {\our} will converge in a finite number of evolution rounds.
\end{theo}
\vspace{-10pt}

\begin{proof}
To prove the above theorem, we only need to prove that the loss term $\mathcal{L}^{(k)}$ for generation $\mathcal{G}^{(k)}$ in {\our} will keep decreasing. Given the $k_{th}$ model generations $\mathcal{G}^{(k)}$, we can denote their introduced loss terms before and after model learning (with Adam) as follows:
\begin{equation}
\mathcal{L}^{(k)} = \sum_{M^{(k)}_i \in \mathcal{G}^{(k)}} \mathcal{L}^{(k)}_i \mbox{, and } \bar{\mathcal{L}}^{(k)} = \sum_{\bar{M}^{(k)}_i \in \bar{\mathcal{G}}^{(k)}} \mathcal{L}^{(k)}_i.
\end{equation}
According to \cite{KB14}, the variables learned by Adam will converge in variable learning, and we have loss term $\bar{\mathcal{L}}^{(k)}$ after learning will be relatively lower than $\mathcal{L}^{(k)}$, i.e., $\bar{\mathcal{L}}^{(k)} \le {\mathcal{L}}^{(k)}$.

Meanwhile, in the model evolution, we know that generation $\mathcal{G}^{(k+1)}$ involves the top $g$ unit models selected from its previous generation $\bar{\mathcal{G}}^{(k+1)} \cup \tilde{\mathcal{G}}^{(k+1)}$. Therefore, we have
\begin{equation}
\mathcal{L}^{(k+1)} \le \bar{\mathcal{L}}^{(k)} \mbox{, where } \mathcal{L}^{(k+1)} = \sum_{M^{(k+1)}_i \in \mathcal{G}^{(k+1)}} \mathcal{L}^{(k+1)}_i.
\end{equation}
In other words, we know $\mathcal{L}^{(k+1)} \le \mathcal{L}^{(k)}$ and the loss term $\mathcal{L}^{(k)}$ will keep decreasing as the number of iterations continues. \QEDB
\end{proof}

\vspace{-14pt}
\subsection{Parallelization of the Learning Algorithm}
\vspace{-8pt}

The training of unit models with {\our} can be effectively deployed on parallel/distributed computing platforms, where each unit model involved in {\our} can be learned with a separate process/server. Among the processes/servers, the communication costs are minor, which exist merely in the \textit{crossover} step. Literally, among all the $g$ unit models in each generation, the communication costs among them is $O(k \cdot g \cdot |\mb{w}|)$, where $|\mb{w}|$ denotes the length of vector $\mb{w}$ and $k$ denotes the required training epochs to achieve convergence by Adam. In the experiment section, we will test the running time of {\our} based on a server with multi-thread CPUs.

%------------------------------------------
\vspace{-10pt}
\section{Numerical Experiments}\label{sec:experiment}
\vspace{-10pt}

To test the effectiveness of {\our}, extensive numerical experiments will be done in this section on several benchmark datasets, including MNIST, ORL, YEAST, ADULT and LETTER. Many existing state-of-the-art baseline methods will be compared with {\our} to show its advantages, and the detailed analyses about convergence, learning settings, time costs will be provided as well.

\vspace{-10pt}
\subsection{LeNet-5 Learning with {\our}}
\vspace{-8pt}

The MNIST dataset involves $60,000$ training instances and $10,000$ testing instances, where each instance is a $28 \times 28$ image with labels denoting their corresponding numbers. Several existing optimization algorithms will be compared with {\our}, which are based on the identical LeNet-5 model architecture introduced in \cite{726791} and a $20\%$ dropout rate is imposed on the fully connection layers in the model. The training batches involve $128$ data instances. In Figure~\ref{fig:train_loss}, we show the introduced loss by different optimization algorithms on the training batch at each learning epoch $[1, 80,000]$ (here, $80,000$ epochs will pass the whole training set for about $170$ rounds). From the results, {\our}, Adam and RMSProp can converge in merely $1,000$ epochs by passing the training set with only $2$ rounds. Meanwhile, via evolution, {\our} can identify a very good starting point (i.e., initial values) and converge much faster than Adam and RMSProp.

In the experiments, Adam, SGD, RMSProp, AdaGrad and AdaDelta are very sensitive to the variable initialization, while {\our} can perform very stably. In Table~\ref{tab:mnist_result}, we show the best evaluation results on record of these optimization algorithms with $80,000$ epochs on the testing set, where the testing instance average loss and the accuracy rate are provided. Besides these optimization algorithm baselines, we also compare {\our} based LeNet-5 with other proposed models in the recent papers at the lower part of Table~\ref{tab:mnist_result}. According to the results, {\our} based LeNet-5 can also outperform these methods with very significant advantages.

%------------------------------------------
\begin{figure*}[t]
\begin{minipage}[b]{0.48\linewidth}
\centering
\subfigure[Parent: Accuracy]{\label{fig:1k_parent_acc}
    \begin{minipage}[l]{0.45\columnwidth}
      \centering
      \includegraphics[width=1.0\textwidth]{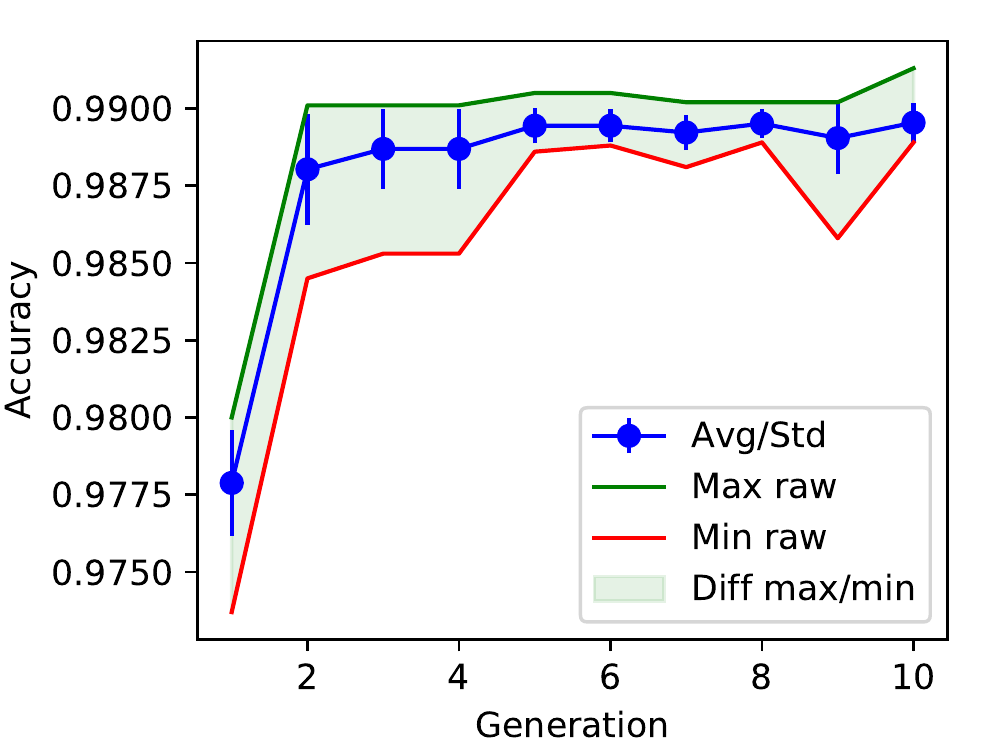}
    \end{minipage}
}
\subfigure[Parent: Loss]{\label{fig:1k_parent_loss}
    \begin{minipage}[l]{0.45\columnwidth}
      \centering
      \includegraphics[width=1.0\textwidth]{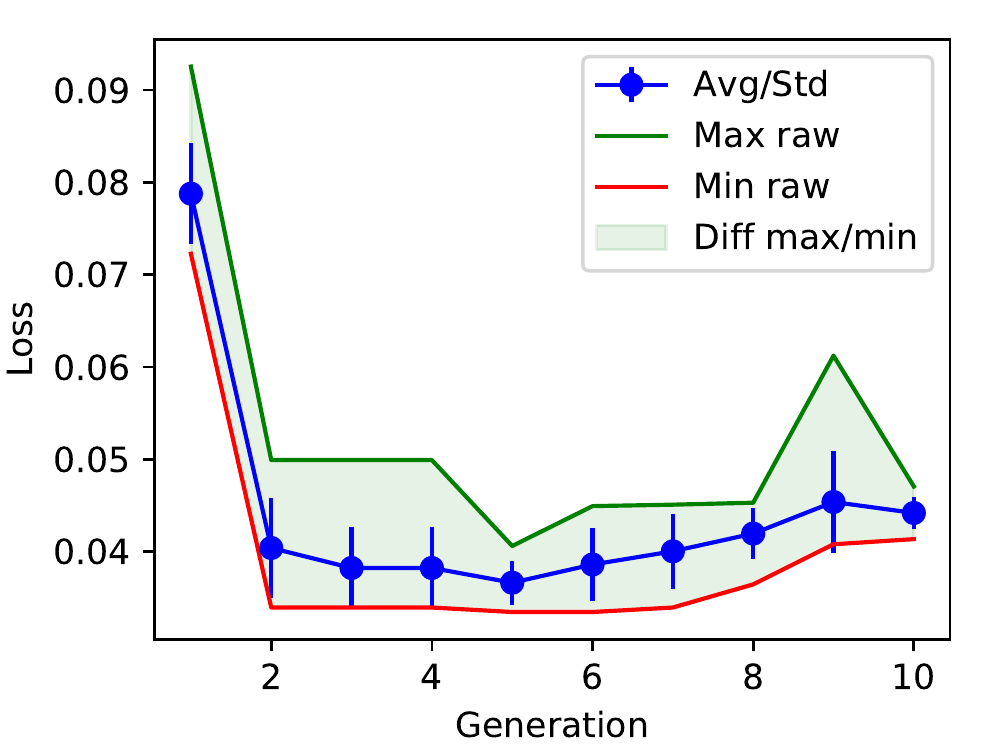}
    \end{minipage}
}
\subfigure[Child: Accuracy]{\label{fig:1k_child_acc}
    \begin{minipage}[l]{0.45\columnwidth}
      \centering
      \includegraphics[width=1.0\textwidth]{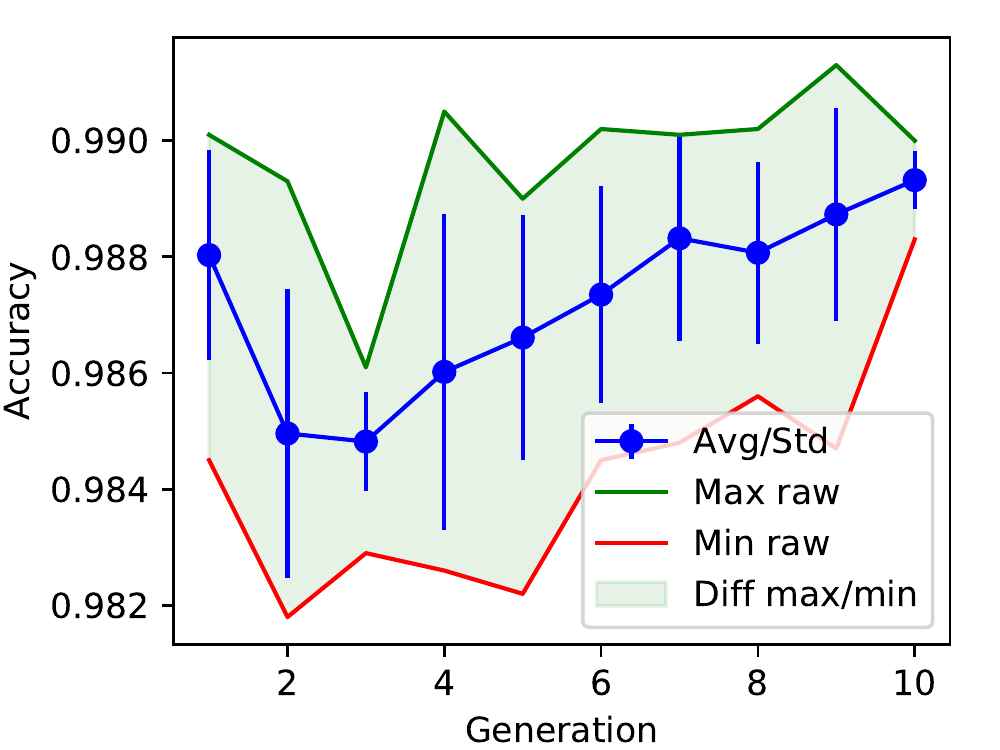}
    \end{minipage}
}
\subfigure[Child: Loss]{\label{fig:1k_child_loss}
    \begin{minipage}[l]{0.45\columnwidth}
      \centering
      \includegraphics[width=1.0\textwidth]{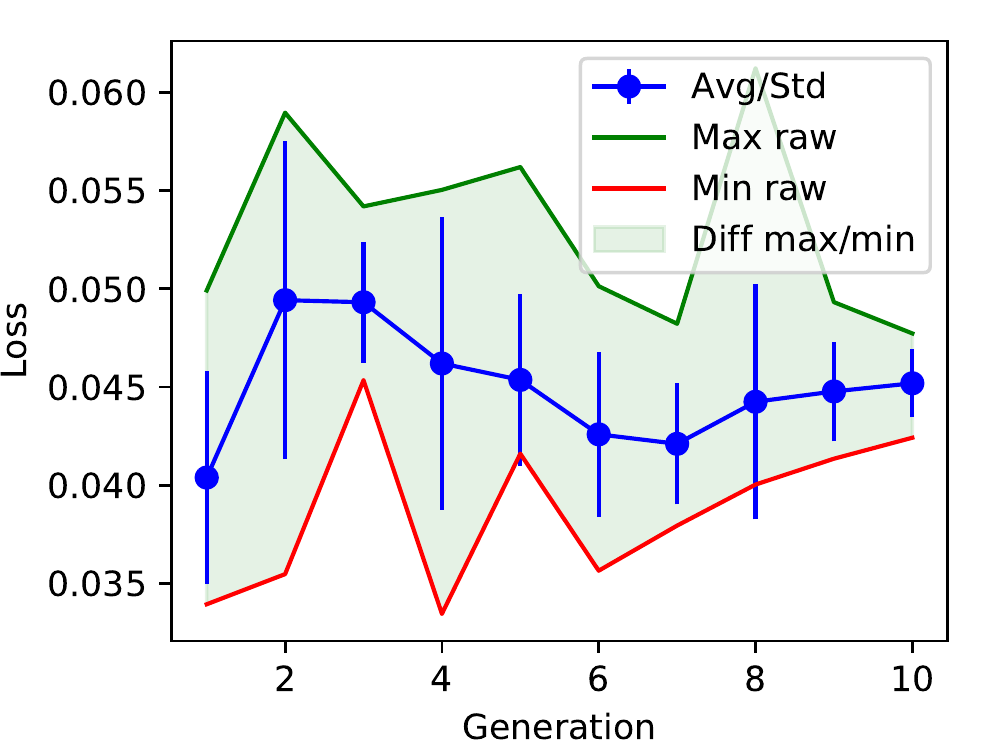}
    \end{minipage}
}
\vspace{-8pt}
\caption{Generation Analysis of {\our}.}\label{fig:sampling}
\end{minipage}\hfill   
\begin{minipage}[b]{0.48\linewidth}
\centering
\subfigure[Parent: Accuracy]{\label{fig:1k_parent_acc}
    \begin{minipage}[l]{0.45\columnwidth}
      \centering
      \includegraphics[width=1.0\textwidth]{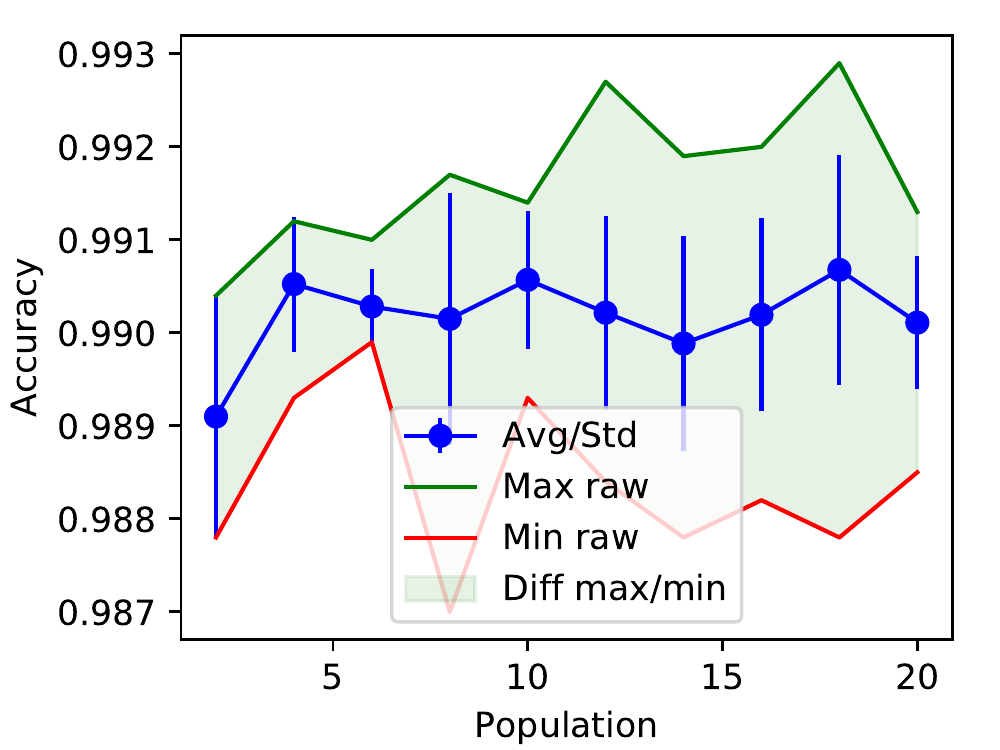}
    \end{minipage}
}
\subfigure[Parent: Loss]{\label{fig:1k_parent_loss}
    \begin{minipage}[l]{0.45\columnwidth}
      \centering
      \includegraphics[width=1.0\textwidth]{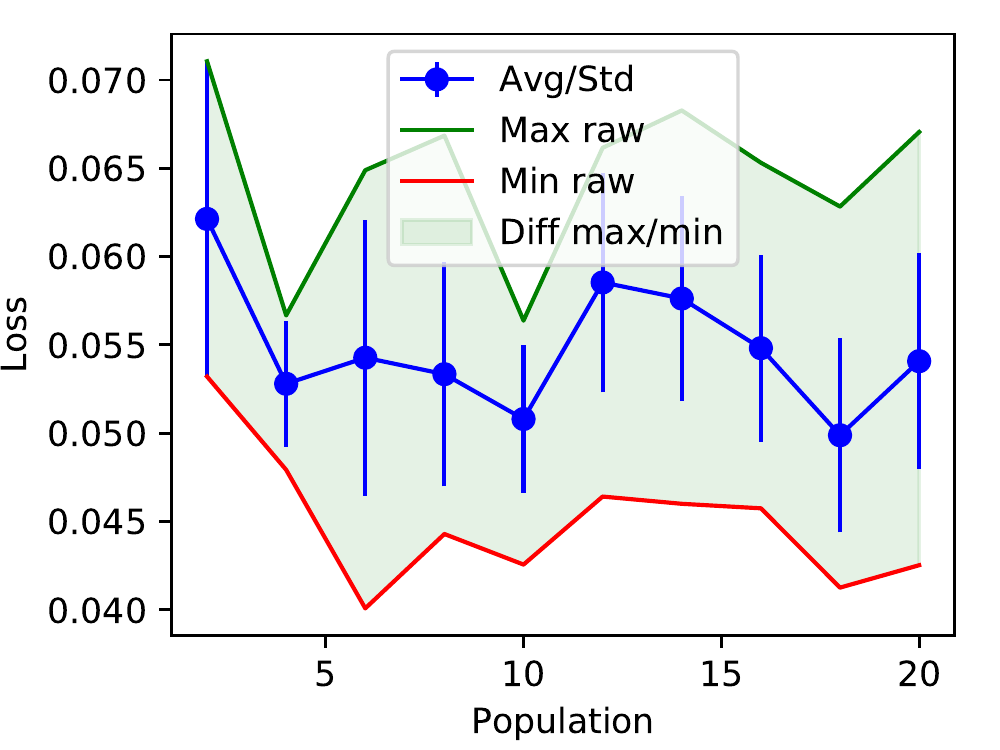}
    \end{minipage}
}
\subfigure[Child: Accuracy]{\label{fig:1k_child_acc}
    \begin{minipage}[l]{0.45\columnwidth}
      \centering
      \includegraphics[width=1.0\textwidth]{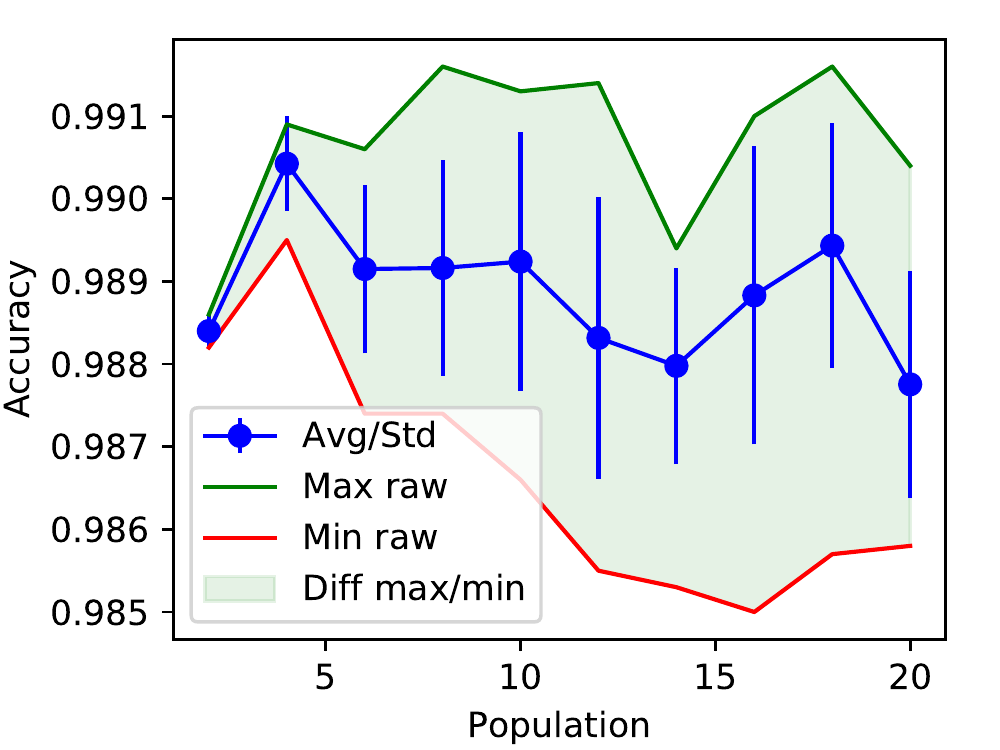}
    \end{minipage}
}
\subfigure[Child: Loss]{\label{fig:1k_child_loss}
    \begin{minipage}[l]{0.45\columnwidth}
      \centering
      \includegraphics[width=1.0\textwidth]{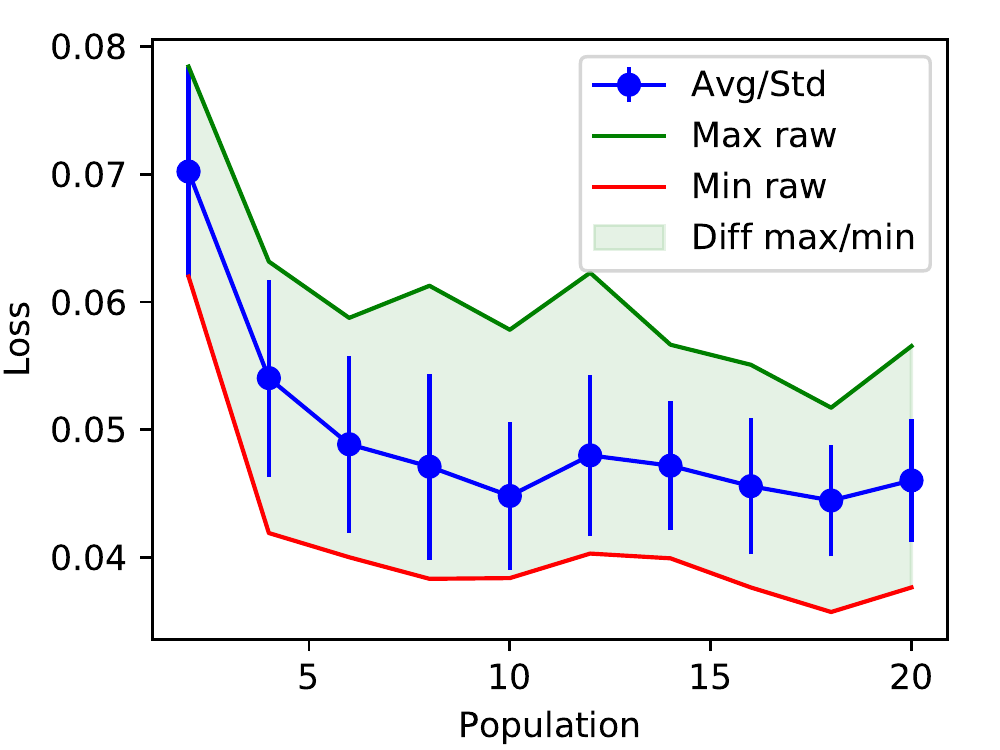}
    \end{minipage}
}
\vspace{-8pt}
\caption{Population Analysis of {\our}.}\label{fig:sampling2}
\end{minipage}
\vspace{-10pt}
\end{figure*}
%------------------------------------------

%----------------
\begin{figure*}[!t]
\begin{minipage}[b]{0.45\linewidth}
 \centering    
 \begin{minipage}[l]{0.9\columnwidth}
  \centering
    \includegraphics[width=1.0\textwidth]{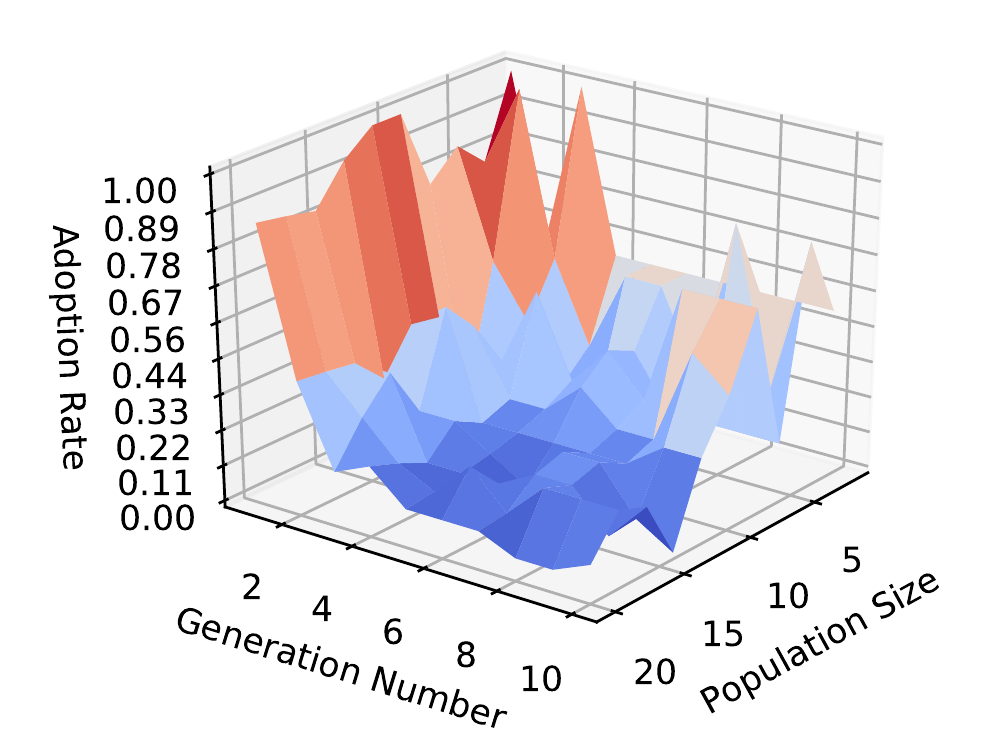}
 \end{minipage}
\caption{Child Model Adoption Rate in Different Settings.}\label{fig:selection}
\end{minipage}\hfill   
\begin{minipage}[b]{0.45\linewidth}
 \centering    
 \begin{minipage}[l]{0.9\columnwidth}
  \centering
    \includegraphics[width=1.0\textwidth]{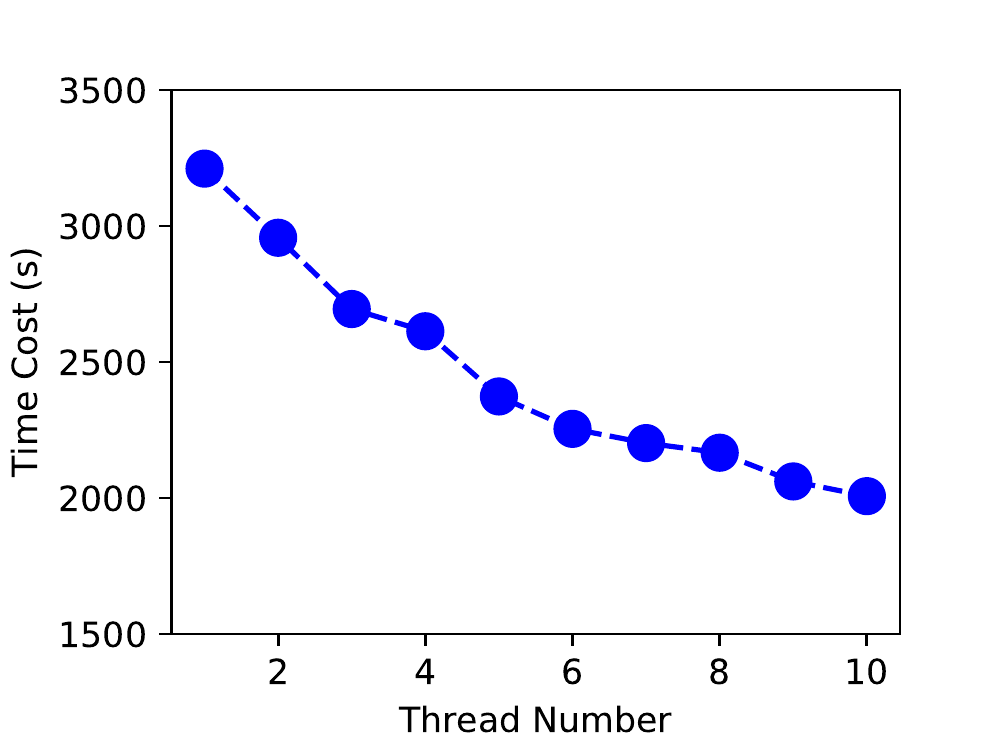}
 \end{minipage}
\caption{Learning Efficiency Analysis of {\our} on Parallel Computing Platform.}\label{fig:efficiency}
\end{minipage}
\vspace{-15pt}
\end{figure*}
%-----------------

%Compared with the other optimization algorithms, {\our} can achieve the best results among the existing optimization algorithms, which obtains an average loss around $0.0396$ and accuracy rate at $99.37\%$. Among the baseline optimization algorithms, Adam and RMSProp can achieve better performance with average loss and accuracy rate at $0.0662$, $99.18\%$ and $0.0745$, $99.00\%$ respectively. The performance of SGD and AdaGrad is slightly under satisfactory, with an average loss around $0.2166$/$0.1720$ and accuracy rate at $94.51\%$/$95.96\%$ respectively. AdaDelta cannot work at all on learning the LeNet-5 model. Besides the optimization algorithm baselines, we also compare {\our} based LeNet-5 with other proposed models in the recent papers at the lower part of Table~\ref{tab:mnist_result}. According to the results, {\our} based LeNet-5 can also outperform these methods with very significant advantages.

\vspace{-10pt}
\subsection{Learning Setting Analysis of {\our}}\label{subsec:settings}
\vspace{-8pt}

In this section, we will provide the analysis of {\our} learning setting parameters, including both the \textit{generation number} and \textit{population size}. In Figure~\ref{fig:sampling}, we provide the analysis about the generation number impact on the performance of {\our} in learning the LeNet-5 model, where the generation number is selected from $\{1, 2, \cdots, 10\}$. According to the results, as the generation number increases, the performance of both the parent and child models will get improved first and then remain stable. It indicates that {\our} can also achieve very fast convergence in generations. In Figure~\ref{fig:sampling2}, we will analyze the affect of the population size parameter on the learning performance of {\our}, where the population size is selected from $\{2, 4, \cdots, 20\}$. According to the results, as a large population of unit models are involved in the {\our} algorithm, the average unit model learning performance of {\our} may vary slightly. Meanwhile, their standard deviation gets larger and larger, and some of the unit models will achieve very outstanding performance, which will be more likely to be selected for evolution to improve the learning performance.

%Meanwhile, with $20K$ training epochs, the learning performance of parent and child models will degrade with more generations, since their accuracy score will decrease while Loss increases as generation number goes up. The potential reason can be the unit models get overfitted during the evolution process. 

%In Figure~\ref{fig:sampling2}, we will analyze the affect of the population size parameter on the learning performance of {\our}, where the population size is selected from $\{2, 4, \cdots, 20\}$. According to the results, as a large population of unit models are involved in the {\our} algorithm, the average unit model learning performance of {\our} may vary slightly. Meanwhile, their standard deviation gets larger and larger, and some of the unit models will achieve very outstanding performance. 

Besides the learning performance of the {\our} algorithm in different settings, we are also interested in the child model adoption rate in the model evolution process, whose results are illustrated in Figure~\ref{fig:selection}. According to the plot, for the {\our} algorithm at the beginning generations, the child models can generally outperform the parent model, and child model adoption rate is very high. Meanwhile, as the generation continues, the parent model will dominate the evolution process. The impact of population on the adoption rate is not significant. As illustrated in the plot, with a larger population, the unit model in {\our} may have various learning performance, while the child model adoption rate is relatively lower. 

%\subfigure[10K-Parent: Accuracy]{\label{fig:10k_parent_acc}
%    \begin{minipage}[l]{0.23\columnwidth}
%      \centering
%      \includegraphics[width=1.0\textwidth]{./acc_10k.pdf}
%    \end{minipage}
%}
%\subfigure[10K-Parent: Loss]{\label{fig:10k_parent_loss}
%    \begin{minipage}[l]{0.23\columnwidth}
%      \centering
%      \includegraphics[width=1.0\textwidth]{./loss_10k.pdf}
%    \end{minipage}
%}
%\subfigure[10K-Child: Accuracy]{\label{fig:10k_child_acc}
%    \begin{minipage}[l]{0.23\columnwidth}
%      \centering
%      \includegraphics[width=1.0\textwidth]{./child_acc_10k.pdf}
%    \end{minipage}
%}
%\subfigure[10K-Child: Loss]{\label{fig:10k_child_loss}
%    \begin{minipage}[l]{0.23\columnwidth}
%      \centering
%      \includegraphics[width=1.0\textwidth]{./child_loss_10k.pdf}
%    \end{minipage}
%}

%---- child model adoption rage in different settings ----

\vspace{-10pt}
\subsection{Efficiency Analysis of {\our}}
\vspace{-8pt}

In Figure~\ref{fig:efficiency}, we study the learning efficiency of {\our} based on the parallel computing setting. Here, the experiments are carried on the Dell T630 EdgePower standalone server with 256 GB memory, 2 Intel E5-2698 CPUs (20 core/40 thread each) and Ubuntu 16.04. Among the generations, the computation is sequential and it is hard to parallelize. Therefore, in the experiments, we just need to test the intra-generation computation efficiency. We set the training epoch number as $5K$ and the unit model population as $10$, but change the number of involved threads with values in $\{1, 2, \cdots, 10\}$. According to the results, as more threads are used in the computing platform, the time cost used by {\our} in learning the LeNet-5 model will decrease first consistently. %For instance, the running time of {\our} with $1$ threads is about $3,211$ seconds, which is about $60\%$ larger than the time cost with $10$ threads.

\vspace{-10pt}
\subsection{Experimental Results of on Other Dataset}
\vspace{-8pt}

In Table~\ref{tab:orl_result}, we show the performance of CNN learned with {\our} with several other baseline methods on the ORL dataset. The ORL dataset \cite{SH94} contains 400 gray-scale facial images of 40 persons. Here, similar to LeNet-5, the CNN model consist of 2 convolutional layers with $16$ and $32$ feature maps of  $5\times5$ kernels, and two 2 $\times$ 2 max-pooling layers. Two full connection layers with $1024$ and $40$ hidden units will connect the convolutional layers with the output layer, where ReLU, cross-entropy and $0.5$-dropout rate are adopted. Here, 5/7/9 images for each person are randomly sampled for model training, and the results on the remaining testing images are provided in Table~\ref{tab:orl_result}. %According to the results, trained with the {\our} algorithm, the CNN model can perform the best among all the baseline methods with great advantages. The accuracy achieved by CNN({\our}) goes up to $100\%$ with $9$ training images for each person. With $5$ training image, CNN({\our}) can also achieve more than $97\%$ accuracy.

%---- ORL CNN ----
% ---- Testing Loss & Accuracy Analysis ---- 
\begin{table}[t]
\begin{minipage}[b]{0.48\linewidth}
\caption{Experiment Results on ORL Dataset.}\label{tab:orl_result}
\scriptsize
\centering
 \begin{tabular}{| l | c | c | c |} 
 \hline
 \multirow{2}{*}{Comparison Methods} &	\multicolumn{3}{c|}{Settings} \\
 \cline{2-4}
  & 5 images & 7 images & 9 images \\
  \hline \hline
 CNN ({\our}) &97.00		&99.17		& 100.00		\\
  \hline
 CNN (Adam) &96.50		&97.50		& 98.50		\\
  \hline
 gcForest & 91.00		& 96.67		& 97.50		\\
   \hline
 Random Forest  & 91.00		& 93.33		& 95.00		\\
  \hline
 SVM (rbf) & 80.50		& 82.50		&85.00		\\
  \hline
 kNN (k=3) & 76.00		& 83.33		&92.50		\\
 \hline
\end{tabular}
%---- ORL CNN ----
% ---- Testing Loss & Accuracy Analysis ---- 
\end{minipage}\hfill   
\begin{minipage}[b]{0.48\linewidth}
\caption{Experiment Results on Other Datasets.}\label{tab:other_result}
\scriptsize
\centering
 \begin{tabular}{| l | c | c | c |} 
 \hline
 \multirow{2}{*}{Comparison Methods} &	\multicolumn{3}{c|}{Datasets} \\
 \cline{2-4}
  & YEAST & ADULT & LETTER \\
  \hline \hline
 MLP ({\our}) &63.70		&87.05		&96.90		\\
  \hline
 MLP (Adam) &62.05 	&85.03		&96.70		\\
  \hline
 gcForest & 63.45		& 86.40		& 97.40		\\
   \hline
 Random Forest & 60.44		& 85.63		& 96.28		\\
  \hline
 SVM (rbf) & 40.76		& 76.41		&97.06		\\
  \hline
 kNN (k=3) & 48.80		& 76.00		&95.23		\\
 \hline
\end{tabular}
\end{minipage}
\vspace{-15pt}
\end{table}
%------------------------------------------
%------------------------------------------

Besides these image datasets and the CNN model, we also carried the empirical experiments on several other types of benchmark datasets and models, whose results are summarized in Table~\ref{tab:other_result}. The datasets used in the experiments include YEAST\footnote{https://archive.ics.uci.edu/ml/datasets/Yeast}, ADULT\footnote{https://archive.ics.uci.edu/ml/datasets/adult} and LETTER\footnote{https://archive.ics.uci.edu/ml/datasets/letter+recognition}, which can be downloaded from the sites at the footnote. Here, we use MLP as the unit model to be learned with {\our}, and we also compare it with several other baseline methods. According to the results, {\our} can also learn better MLP model than Adam. Furthermore, compared with the other baseline methods, MLP ({\our}) can effectively outperform them except the LETTER dataset, where gcForest performs slightly better than MLP ({\our}).

%------------------------------------------
\vspace{-10pt}
\section{Conclusion}\label{sec:conclusion}
\vspace{-10pt}

In this paper, we have introduced a novel optimization algorithm, i.e., {\our}, which is capable to learn the variables for deep learning models both effectively and efficiently. By combining Adam and genetic algorithm together, {\our} can learn the unit models with Adam and evolve them with the genetic algorithm. At the same time, {\our} also integrates the advantages of Adam and genetic algorithm, which can converge fast, avoid sinking into the local optima, and be effectively deployable on parallel/distributed computing platforms. Extensive experiments have been done on many real-world benchmark datasets, and the experimental results show that {\our} can outperform the existing optimization algorithms in deep neural network learning in terms of both its effectiveness and efficiency.
%------------------------------------------

\bibliographystyle{abbrv}
\bibliography{reference}
%------------------------------------------

\end{document}